\documentclass[runningheads]{llncs}
\usepackage[T1]{fontenc}
\usepackage{graphicx}
\usepackage{hyperref}
\usepackage{color}

\usepackage{times}
\usepackage{epsfig}
\usepackage{amsmath}
\usepackage{amssymb}
\usepackage{bm}
\usepackage{booktabs}
\usepackage{tablefootnote}
\usepackage{color}
\usepackage{multirow}

\newcommand{\CR}[1]{#1}

\begin{document}

\title{Gigapixel Whole-Slide Images Classification using Locally Supervised Learning}
%
\newcommand*\samethanks[1][\value{footnote}]{\footnotemark[#1]}

\author{Jingwei Zhang\inst{1}\thanks{These authors contributed equally to this paper.} \and
Xin Zhang\inst{1}\samethanks[1] \and Ke Ma \inst{2} \and Rajarsi Gupta\inst{1} \and
Joel Saltz\inst{1} \and Maria Vakalopoulou\inst{3} 
\and Dimitris Samaras\inst{1}}

%
\authorrunning{J. Zhang et al.}
\institute{Stony Brook University, USA \and Snap Inc., USA
\and
CentraleSupélec, University of Paris-Saclay, France\\
{ \email{\{jingwezhang, xin.zhang, kemma, samaras\}@cs.stonybrook.edu}} \\
\email{\{Rajarsi.Gupta, Joel.Saltz\}@stonybrookmedicine.edu   maria.vakalopoulou@centralesupelec.fr} }


%
\maketitle              
\begin{abstract}
Histopathology whole slide images (WSIs)  play a very important role in clinical studies and serve as the gold standard for many cancer diagnoses.
However, generating automatic tools for processing WSIs is challenging due to their enormous sizes.
Currently, to deal with this issue, conventional methods rely on a multiple instance learning (MIL) strategy to process a WSI at patch level. 
Although effective, such methods are computationally expensive, because tiling a WSI into patches takes time and does not explore the spatial relations between these tiles.
To tackle these limitations, we propose a locally supervised learning framework which processes the entire slide by exploring the entire local and global information that it contains.
This framework divides a pre-trained network into several modules and optimizes each module locally using an auxiliary model. We also introduce a random feature reconstruction unit (RFR) to preserve distinguishing features during training and improve the performance of our method by $1\%$ to $3\%$.
Extensive experiments on three publicly available WSI datasets: TCGA-NSCLC, TCGA-RCC and LKS, highlight the superiority of our method on different classification tasks. 
Our method outperforms the state-of-the-art MIL methods by $2\%$ to $5\%$ in accuracy, while being $7$ to $10$ times faster. Additionally, when dividing it into eight modules, our method requires as little as 20\% of the total gpu memory required by end-to-end training. Our code is available at \href{https://github.com/cvlab-stonybrook/local_learning_wsi}{https://github.com/cvlab-stonybrook/local\_learning\_wsi}

\end{abstract}
\section{Introduction}

Computational pathology involving observation of tissue slides with a microscope, is the gold standard for cancer diagnosis. In recent years, digital pathology has emerged as a powerful technology for digitizing whole slide images (WSIs) for assessment, sharing and analysis \cite{dimitriou2019deep}. This provides researchers a good opportunity to develop computer-aided analysis systems for various levels of applications, such as cell counting, gland segmentation, and WSI classification~\cite{deng2020deep,hou_robust_2019,lerousseau2020weakly}. In particular, WSI-based cancer diagnosis faces unique challenges. The most typical characteristics of WSIs are their extremely large image size and high resolution. A WSI generally can be as large as 100,000$\times$100,000 pixels at a $40$X magnification, which makes it impractical to train deep neural networks in an end-to-end (E2E) manner. Consequently, the most popular methods nowadays follow a patch-based paradigm \cite{le2020utilizing_breast,patch_coudray2018classification_nat_med}, i.e. each WSI is first tiled into thousands of small patches.
Then a model extracts and aggregates patch-level features to make the final prediction \cite{li2021dsmil,shao2021transmil,whole_wsi_by_patch_compression_tellez2019neural}. 

Such methods follow a Multiple Instance Learning (MIL) scheme which is currently the state-of-the-art for solving histopathology classification tasks~\cite{Hou_2016_CVPR,campanella2019milrnn,shao2021transmil,zhang2021joint,li2021dsmil}. Zhang et al. \cite{zhang2021joint} proposed a spatial and magnification based attention sampling strategy to extract informative patches, and directly learned a WSI classification model on these patches. DSMIL~\cite{li2021dsmil} jointly trained a patch and an image classifier, where the patches are selected softly with  instance-level attention. 
More recently,
TransMIL~\cite{shao2021transmil} presented a transformer-based MIL framework to explore both morphological and spatial information among instances. 
However, such a technical paradigm has some intrinsic shortcomings; these methods do not explore the spatial relations of each tile, by failing to properly combine the local and global information of the tumor's microenvironment. Moreover, these methods rely on pretrained features to represent the tiles since the current deep learning architecture cannot be trained in an end-to-end manner.

The end-to-end training of deep neural networks requires storing in memory the entire computational graph as well as the layer activations during the forward pass. Then the loss backpropagates and updates the weights layer by layer based on the chain rule. Storage of the graph and the gradients occupy a large amount of GPU memory, limiting the input image size. 
Some researchers \cite{takahama2019multi,pinckaers2020streaming} proposed to retain gradient information and train the model part by part to reduce memory consumption. Nevertheless, they still tiled images into patches, and trained networks on smaller regions with limited receptive field sizes.

Due to E2E training's limited scalability to large input and large architectures, recent research attempts to seek alternatives to mitigate the memory constraints, among which locally supervised learning attracts increasing interest~\cite{belilovsky2020decoupled,wang2020revisiting}. Locally supervised learning aims to train each layer locally with a pre-defined objective function, without backpropagating the gradients end-to-end. The network training is free from storing \textsl{all} intermediate variables and the memory consumption is thus reduced. 
Belilovsky et al. \cite{belilovsky2019greedy} attached an auxiliary convolutional neural network classifier at each local module to predict the final target and evaluated it on ImageNet \cite{deng2009imagenet}.
N{\o}kland et al. \cite{pmlr-v97-nokland19a} proposed to use both classification loss and contrastive loss to supervise each local module and showed it was better than using a single loss.

In this paper, we introduce a locally supervised learning paradigm to train a classification network using the entire WSI. Our method splits a deep network into multiple gradient-isolated modules and each part is trained separately with local supervision.
Thus we can use the entire WSI as input and do not have a patch size limited receptive field.
Moreover, we further propose the Random Feature Reconstruction (RFR) model to boost the performance and optimize the GPU usage. To the best of our knowledge, we are the first to propose a locally supervised learning scheme coupled with RFR for the classification of entire WSIs. Our method has been extensively evaluated on three public WSI datasets, and achieves state-of-the-art performance compared to MIL-based methods. Moreover, without tiling, our method is significantly faster during inference.

\section{Method}

\begin{figure}[ht]
\begin{center}
\includegraphics[width=\linewidth]{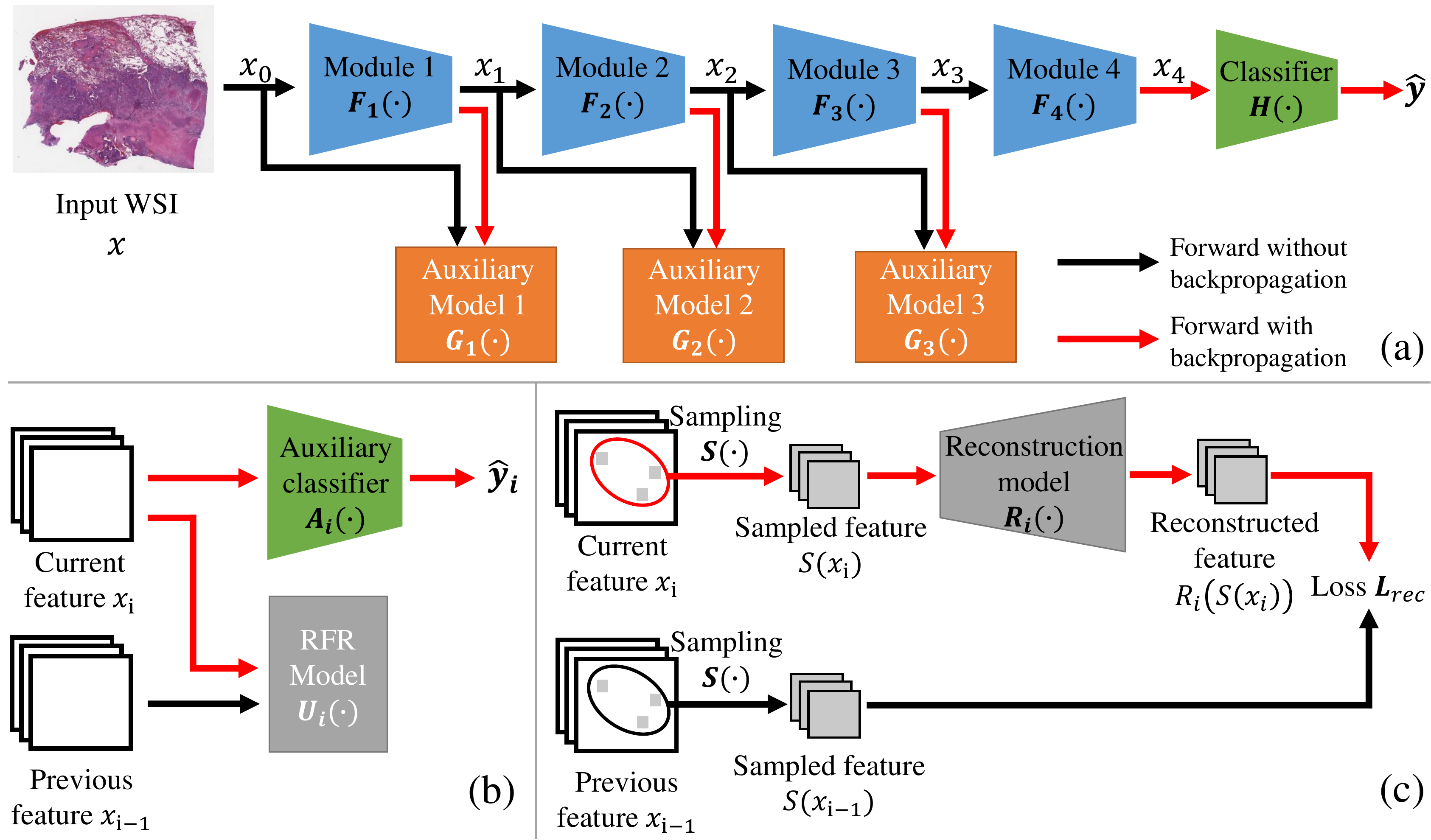}
\end{center}
   \caption{Overview of the proposed method.
   (a) Overall structure of our locally supervised learning method. 
   A network is divided into 4 modules and a classifier. 
   The first 3 modules $F_i, i=1, 2, 3$ are optimized using auxiliary models $G_i, i=1, 2, 3$ respectively.
   The last module $F_4$ is optimized together with the classifier $H(\cdot)$.
   We assume $x_0 = x$.
   (b) Structure of the auxiliary model $G_i(\cdot)$. 
   It has an auxiliary classifier $A_i(\cdot)$ and a Random Feature Reconstruction (RFR) model $U_i(\cdot)$.
   (c) Structure of the RFR model. It reconstructs randomly sampled regions in the previous feature map.
   }
\label{fig:framework}
\end{figure}

The key idea of our locally supervised learning is dividing a network layer by layer into several consecutive modules and optimizing them separately.
Formally, let us assume without loss of generality, a network $F(\cdot)$ composed by $K$ consecutive modules:
$F(\cdot) = ((H \circ F_K) \circ F_{K-1} \circ \dots \circ F_1)(\cdot)$, where $F_i(\cdot)$ represents the $i$-th network module. $H(\cdot)$ is a gated attention multiple instance learning\cite{ilse2018abmil} based classifier and
$\circ$ is the function composition operation. Such a network is trained using pairs of $(x, y)$ on which $x$ denotes the entire WSI and $y$ the corresponding label. 
 
A network module contains several network layers of the original network, for example, the first 6 layers in a ResNet34.
The input to a network module $F_i, i=1,..,K$ is $x_{i - 1}$ and the output is $x_i$, assuming $x_0 = x$.
An overview of our approach is presented in Fig.\ref{fig:framework} on which the forward and backward passes are indicated.

In such a setup, each module is trained locally. More specifically, given a network module $F_i$ and its input $x_{i - 1}$, we use an auxiliary model $G_i$, and compute the loss as $\mathcal{L}_i = G_i(F_i(x_{i - 1}), x_{i-1}, y) = G_i(x_i, x_{i - 1}, y)$.
We train $F_i$ by minimizing the $\mathcal{L}_i$.
Then, the trained module $F_i$ is frozen and in an iterative process, the same technique is applied to $F_{i + 1}$ by minimizing $\mathcal{L}_{i + 1} = G_{i + 1}(x_{i + 1}, x_i, y)$. The same process is applied to each of the $K-1$ modules. Finally, the  final module $F_K(\cdot)$ is optimized together with the classifier $H(\cdot)$ without an additional auxiliary model since the label $y$ serves as the final supervision.

\subsection{Auxiliary model}

The training of each model is performed using an auxiliary model with a greedy strategy~\cite{belilovsky2019greedy}.
As shown in Fig.\ref{fig:framework}(b), the auxiliary model has two parts: 
an auxiliary classifier $A_i(\cdot)$ and an RFR model $U_i(\cdot)$. 
The auxiliary classifier has a similar structure of classifier $H(\cdot)$ and computes a classification loss $\mathcal{L}_{cls}(\hat{y}_i, y)$ between the prediction $\hat{y}_i = A_i(x_i)$ and ground truth $y$. Such a design enables the training of the module $F_i(\cdot)$ locally. 
However, as discussed in~\cite{wang2020revisiting}, the shallower layers in a network have limited ability to extract discriminative features, making the training difficult. 

To overcome this problem, the authors proposed to reconstruct the input image $x$ from the feature map $x_i$ and applied a reconstruction loss as a regularization to preserve the discriminative features.
However, this strategy cannot be applied to WSIs as reconstructing an entire WSI is too costly. 

To deal with this issue, we propose to use a RFR model instead of the reconstruction module . More specifically, a RFR model reconstructs randomly sampled regions from the previous feature map.

As shown in Fig.\ref{fig:framework}(c), the first step of a RFR model $U_i(\cdot)$ is to randomly sample ($S(\cdot)$), $10$ corresponding spatial locations on the latent representations from the $i$-th module $x_i$ as well as from the previous module $x_{i-1}$.
Feature patches $S(x_i)$ and $S(x_{i-1})$ are cropped according to the sampled spatial locations.
Then, a reconstruction network $R_i(\cdot)$ is applied to the cropped features from $i$-th module $S(x_i)$, aiming to reconstruct the target $S(x_{i-1})$.
A reconstruction loss $\mathcal{L}_{rec}$ is used to minimize the distance between the reconstructed feature patches $R_i(S(x_i))$ and feature patches from its previous module in the corresponding spatial locations $S(x_{i-1})$.
During training, this random sampling process eventually iterates over most locations and thus encourages the network to preserve discriminative features with limited GPU memory cost.

\subsection{Optimization}
In our framework, the first $K-1$ modules are optimized locally with the following setting:
\begin{align}
    \mathcal{L}_i &= \mathcal{L}_{cls}(A_i(F_i(x_{i-1})), y) + \alpha \mathcal{L}_{rec}(R_i(S(F_i(x_{i-1}))), S(x_{i-1})),
    \\
    i&=1,\dots, K-1,
\end{align}
where $x_{i-1}$ is the input feature of the $i$-th module $F_i(\cdot)$ and hyperparameter $\alpha$ is a regularization term. Since the last module $F_K(\cdot)$ is jointly optimized with the  classifier $H(\cdot)$ the training scheme is changed to:
\begin{align}
    \mathcal{L}_K &=\mathcal{L}_{cls}(H(F_K(x_{K-1})), y),
\end{align}
where $x_{K-1}$ is the input feature of the last module.
We used L1 loss as the reconstruction loss $\mathcal{L}_{rec}$ and cross entropy loss as the classification loss $\mathcal{L}_{cls}$. We set $\alpha$ to be $1$ after grid search on the validation dataset. However, our methodological design is independent of these losses and different reconstruction and classification losses can be applied.

\section{Experiments and Discussion}
\subsection{Datasets}

\subsubsection{TCGA-NSCLC} The TCGA-NSCLC (The Cancer Genome Atlas-Non-Small Cell Lung Cancer) dataset includes two sub-types of lung cancer, Lung Adenocarcinoma (LUAD) and Lung Squamous Cell Carcinoma (LUSC).
The dataset contains a total of 1053 diagnostic WSIs. We randomly split them into 663 training slides, 166 validation slides and 214 testing slides (10 slides without magnification labels are discarded). We benchmarked the performance of our model on this dataset for the lung cancer sub-type classification task. The WSIs were on $5$X magnification and the size of the slides ranges from $1581\times1445$ to $23362\times11345$.

\subsubsection{TCGA-RCC} The TCGA-RCC (Renal Cell Carcinoma) dataset includes three sub-types of kidney cancer, Kidney Chromophobe Renal Cell Carcinoma (KICH), Kidney Renal Clear Cell Carcinoma (KIRC) and Kidney Renal Papillary Cell Carcinoma (KIRP).
The dataset contains a total of 939 diagnostic digital slides.
We randomly split them into 603 training slides, 150 validation slides and 186 testing slides. 
We benchmarked the performance of our model on this dataset for the classification of these three different kidney cancer types. The WSIs were on $5$X magnification and the size of the slides ranges from $2610\times1351$ to $23849\times10257$.

\subsubsection{LKS} The Liver-Kidney-Stomach(LKS)~\cite{maksoud2020sos} dataset is a multi-tissue indirect immuno-fluorescence slides dataset. 
It includes four classes: Negative, Anti-Mitochondrial Antibodies (AMA), Vessel-Type Anti-Smooth Muscle Antibodies (SMA-V) and Tubule-Type Anti-Smooth Muscle Antibodies (SMA-T). 
The dataset contains a total of 684 slides, including 205 testing slides. To perform our experiments, we further split the rest into 383 training slides and 96 validation slides.
Each slide in this dataset has the original size of $40000 \times 40000$. 
We further resized the images to $10000 \times 10000$ at 5X magnification.

\subsection{Implementation Details}
For all our experiments, we used ResNet34\cite{he2016deep} pretrained on ImageNet\cite{deng2009imagenet} as our backbone network ($F(\cdot)$). 
We froze the first 4 layers of ResNet34 and increased the stride of the first convolution from 2 to 3 to enlarge the receptive field. 
\CR{We set the batch size to be $1$, since each WSI has a different size. To mitigate potential instability, we used the common optimization practice of accumulating the gradients of 8 batches before updating the parameters.}
All the batch normalization layers were frozen as well. 
In the RFR model, we sampled $10$ patches of the size $128\times128$ for the first auxiliary model $G_1(\cdot)$ and the spatial dimensions sampled in the following modules depended on the size of the feature map.
The number and size of patches are determined by the validation dataset.
We used a gated attention multiple instance learning (GABMIL)\cite{ilse2018abmil} based network in the auxiliary classifier $A_i(\cdot)$ and classifier $H(\cdot)$.

We used AdamW\cite{loshchilov2017adamW} with weight decay $10^{-6}$ as the optimizer. 
For the two TCGA datasets, the learning rate was initially set to $1\times10^{-5}$ for the pre-trained backbone modules $F_i(\cdot)$ and $2\times10^{-5}$ for the randomly initialized auxiliary models $G_i(\cdot)$ and classifier $H(\cdot)$. Learning rates were decreased by a factor of 0.1 when the loss and validation accuracy plateaued. 
For the LKS dataset, the initial learning rates were doubled to $2\times10^{-5}$ for pre-trained backbone modules and $4\times10^{-5}$ for randomly initialized auxiliary models. 

We used the PyTorch library\cite{paszke2019pytorch} and trained our models on 
a NVIDIA Tesla V100 or a Nvidia Quadro RTX 8000 GPU.

\begin{table}[b!]
\caption{Comparison of accuracy and AUROC on three datasets. Our method, of both $K=4$ and $K=8$, outperforms existing state-of-art MIL models}
\label{table:result:accuracy}
\begin{center}
\setlength{\tabcolsep}{1.6mm}{

\begin{tabular}{c r r r r r r}
\toprule
\multicolumn{1}{c}{Dataset} & \multicolumn{2}{c}{TCGA-NSCLC} & \multicolumn{2}{c}{TCGA-RCC} &
\multicolumn{2}{c}{LKS}\\
Metric & \multicolumn{1}{c}{Accuracy} & \multicolumn{1}{c}{AUROC} & \multicolumn{1}{c}{Accuracy} & \multicolumn{1}{c}{AUROC} & \multicolumn{1}{c}{Accuracy} & \multicolumn{1}{c}{AUROC} \\
\midrule
Max-pooling 
    & 0.8318 & 0.9036 
        & 0.8495 & 0.9306 
            & 0.8049 & 0.9366 \\ 
Avg-pooling 
    & 0.7944 & 0.8669
        & 0.8172 & 0.9309 
            & 0.6000 & 0.9086  \\ 
ABMIL~\cite{ilse2018abmil}
    & 0.8037 & 0.8816 
        & 0.8495 & 0.9423
            & 0.8341 & 0.9392 \\ 
GABMIL~\cite{ilse2018abmil}
    & 0.8364 & 0.8762 
        & 0.8602 & 0.9535 
            & 0.8146 & 0.9399 \\ 
MIL-RNN~\cite{campanella2019milrnn}
    & 0.8178 & 0.9011 
        & \multicolumn{1}{c}{/} &  \multicolumn{1}{c}{/} 
            & \multicolumn{1}{c}{/} & \multicolumn{1}{c}{/} \\ 
DSMIL~\cite{li2021dsmil}
    & 0.8271 & 0.8909
        & 0.8710 & 0.9590  
            & 0.8390 & 0.9328 \\ 
CLAM-SB~\cite{lu2021clam}
    & 0.8224 & 0.9185 
        & 0.8763 & 0.9701 
            & 0.8293 & 0.9446 \\
CLAM-MB~\cite{lu2021clam}
    & 0.8598 & 0.9131 
        & 0.8763 & 0.9716
            & 0.8439 & 0.9448 \\
\midrule
StreamingCNN~\cite{pinckaers2020streaming} 
    & 0.8692          & 0.9260       
        & 0.8817         & 0.9660      
            & 0.8927      & \textbf{0.9652}    \\ 
\midrule
Ours (K=4) 
    & \textbf{0.8785} & \textbf{0.9377} 
        & \textbf{0.9140} & 0.9740 
            & \textbf{0.8976} & \textbf{0.9562}\\ 
Ours (K=8) 
    & \textbf{0.8785} & 0.9322 
        & 0.9032 & \textbf{0.9760}
            & 0.8829 & 0.9633 \\
\bottomrule
\end{tabular}
}
\end{center}
\end{table}

\subsection{Results}

\subsubsection{Evaluation of overall performance}

We chose overall accuracy and area under Receiver Operating Characteristic curve (AUROC) as the main metrics to evaluate our method. 
Our baselines included ImageNet pre-trained ResNet34 with two different pooling methods: average pooling and max pooling. 
We also included the current state-of-the-art deep MIL models: the attention based multiple instance learning (ABMIL)\cite{ilse2018abmil} and its gated variant GABMIL~\cite{ilse2018abmil}, dual stream attention based model DSMIL~\cite{li2021dsmil}, single-attention-branch CLAM-SB~\cite{lu2021clam}, multi-attention-branch CLAM-MB~\cite{lu2021clam},
, and also two-stage recurrent neural network based aggregation MIL-RNN~\cite{campanella2019milrnn}, which considers binary classifications only.
\CR{All these baselines are trained on 5X resolution and using ResNet34 for fair comparision.}
Our method was able to fine tune the ImageNet pretrained weights to adapt to the medical image domain, while other methods directly used the ImageNet pretrained features.

As shown in Table~\ref{table:result:accuracy}, our method ($K=4$) outperformed all the compared methods in the overall accuracy and AUROC metrics. 
To the best of our knowledge the SOTA for the TCGA-NSCLC dataset is $96.3\%$ AUROC ($95\%$ confidence interval: $93.7\%$–$99.0\%$) reported by CLAM \cite{lu2021clam}. For experimental uniformity, we used the exact same splits on all comparisons and reported accuracy and AUROC, reporting a $91.9\%$ AUROC for CLAM for the same resolution which is also higher than the rest of the compared methods. Our method achieved $1.87\%$ higher accuracy than the best compared method CLAM-MB, and $2.46\%$ higher AUROC.
On the TCGA-RCC dataset, our method achieved $3.77\%$ higher accuracy and $0.24\%$ higher AUROC.
On the LKS dataset, our method had $5.37\%$ higher accuracy and $1.14\%$ higher AUROC compared with the best performing method CLAM-MB. 
\CR{
Also, comparing with the SOTA on this dataset, SOS\cite{maksoud2020sos}, our method achieved comparable performance to it ($90.73\%$ accuracy).
}
 
Moreover, Table~\ref{table:result:accuracy} highlights the robustness of our method with respect to the different modules. In particular, our model was divided into $8$ modules and each of them trained locally using the proposed strategy performs as well as the $K=4$ and outperformed the compared methods.
\CR{
Also, we compared our method with a non-MIL approach StreamingCNN~\cite{pinckaers2020streaming}, as shown in Table~\ref{table:result:accuracy}, our method ($K=4$) outperformed it on TCGA-NSCLC and TCGA-RCC. On LKS dataset, our method achieved  better accuracy and comparable AUROC.
}

\begin{table}[t!]
\caption{Comparison of GPU memory consumption. Our method required around $20\%$($K=8$) to $30\%$($K=4$) GPU memory compared to E2E training. * is measured on CPU because of GPU memory limitation}
\begin{center}
\begin{tabular}{lccc}
\toprule
Image size & $8698\times7496$ & $12223\times10057$ & $23849\times10257$ \\
\midrule
E2E        & 17.89G           & 33.63G            & 78.14G*            \\
Ours (K=4) & 5.35G            & 9.64G             & 18.32G\text{   }             \\
Ours (K=8) & 3.71G            & 6.47G             & 11.85G\text{   }            \\
\bottomrule
\end{tabular}
\end{center}
\label{table:result:gpu}
\end{table}

\begin{table}[t!]
\caption{Comparison of inference speed on three different sized inputs. Our method ran $7$ to $10$ times faster than GABMIL since  we do not have the tilling and feature extraction step.}
\label{table:result:speed}
\begin{center}
\begin{tabular}{lccccccccc}
\toprule
Image size
    & \multicolumn{3}{c}{$8698\times7496$}
        & \multicolumn{3}{c}{$12223\times10057$}
            & \multicolumn{3}{c}{$23849\times10257$}                        \\
Method
    & GABMIL  & Ours   & Speed gain 
        & GABMIL & Ours & Speed gain 
            & GABMIL & Ours & Speed gain \\
\midrule
Tiling
    & 0.3s    & /    &     /       
        & 0.7s    & /     &    /          
            & 1.8s     & /    &     /     \\
Features 
    & 2.6s   & /      &    /     
        & 3.7s   & /      &   /        
            & 9.6s    & /     &   /         \\
Prediction          
    & <0.1s  & 0.3s   &   /        
        & <0.1s & 0.6s   &   /         
            & <0.1s & 1.2s   &  /          \\
Total
    & 2.9s & \textbf{0.3}s &  9.7x       
        & 4.4s & \textbf{0.6}s & 7.3x       
            & 11.4s & \textbf{1.2}s & 9.5x \\
\bottomrule
\end{tabular}
\end{center}
\end{table}

\subsubsection{Evaluation of GPU memory consumption}
Another major advantage of our method is that our method significantly reduced the GPU memory required during training and thus enables training on the entire WSI. 
We compared the GPU memory consumption of our method (for $K=4$ and $K=8$) and that of the end-to-end (E2E) training.
As the sizes of images in a WSI dataset usually vary a lot, instead of evaluating the memory consumption on three datasets, we evaluated it on three different sized images: a $23849 \times 10257$ image, the largest image in the TCGA-RCC dataset, a $12223\times10057$ image, and an $8698\times7496$ image.
Note that for the $23848\times10257$  image, we performed the E2E measurement on the CPU since the GPU memory was not enough to perform this task.
As shown in Table~\ref{table:result:gpu}, when the input image size was $8698\times7496$, our 4 divided network required only $29.9\%$ of the GPU memory that the E2E training needs.
This number further dropped to $20.7\%$ if we divided the network into $K=8$ modules.

The same memory usage held for the other two input image sizes. In general, our $4$ divided network required only around $30\%$ memory and our $8$ divided network required only around $20\%$ memory compared to E2E training.

\subsubsection{Evaluation of time efficiency}

Besides the higher accuracy and the lower GPU memory cost, our method is faster in inference than the standard MIL approaches. 
We measured the total time (in seconds) that our method requires to infer an entire WSI and compared it with GABMIL~\cite{ilse2018abmil}, a high performance MIL model. We timed the whole inference pipeline including patch tiling, feature extraction, and final prediction.
We reported the total inference time on WSIs in 3 different sizes.
Table~\ref{table:result:speed} highlights the time efficiency of our method. Our method took only $1.2$ seconds to classify a 23849$\times$10257 WSI, while GABMIL needed more than $11.4$ seconds due to the time consuming step of feature extraction on the large amount of patches.

\subsubsection{Ablation Study on RFR}

We conducted an ablation study on the Random Feature Reconstruction (RFR) model. 
Table~\ref{table:result:rfr} shows the comparison on the accuracy of our method with and without RFR.
On the TCGA-NSCLC dataset, using the RFR model improved the accuracy by $0.47\%$ for $K=4$ and $3.27\%$ for $K=8$.
On the TCGA-RCC dataset, using the RFR model improved the accuracy by $2.15\%$ for $K=4$ and $3.22\%$ for $K=8$.
On the LKS dataset, using the RFR model improved the accuracy by around $1\%$ for both $K$=4 and $K$=8.

\begin{table}[t!]
\caption{Comparison of accuracy of our method with and without RFR. The RFR model improved the accuracy of our method by $1\%$ to $3\%$.}
\begin{center}
\begin{tabular}{lccc}
\toprule
Dataset       & TCGA-NSCLC & TCGA-RCC & \CR{LKS}    \\
\midrule
K=4, w/o RFR &  0.8738          & 0.8925            & 0.8829 \\
K=4, w. RFR   & \textbf{0.8785} & \textbf{0.9140}   & \textbf{0.8976} \\
\midrule
K=8, w/o RFR  &  0.8458         & 0.8710           & 0.8780 \\
K=8, w. RFR   & \textbf{0.8785} & \textbf{0.9032}   & \textbf{0.8829} \\
\bottomrule
\end{tabular}
\end{center}
\label{table:result:rfr}
\end{table}
\section{Conclusion}
In this paper, we introduced a locally supervised learning framework to train using entire whole slide images.
We evaluated it on three WSI datasets and achieved a $2\%$ to $5\%$ accuracy improvement compared to existing MIL methods.
This significant performance gain was achieved by reducing GPU memory consumption and enabling fine-tuning of the feature extractor. 
Compared with end-to-end training, our method required only $20\%$ to $30\%$ of the memory.
Moreover, our method did not require tiling as existing MIL methods do, thus it was 7 to 10 times faster during inference.
We also demonstrated that the proposed random feature reconstruction (RFR) model improved the performance of our locally supervised learning framework by $1\%$ to $3\%$.
Our proposed approach showed the greater potential of locally supervised learning on classifying whole slide images and we will explore its applications on other tasks including segmentation.

\subsubsection{Acknowledgements} This work was partially supported by the ANR Hagnodice ANR-21-CE45-0007, the NSF IIS-2123920 award, Stony Brook Cancer Center donors Bob Beals and Betsy Barton  as well as the Partner University Fund 4D Vision award.

%
%
%
\bibliographystyle{splncs04}
\bibliography{bibliography.bib}
\end{document}